\documentclass[conference]{IEEEtran}
\usepackage{times}

\usepackage[numbers]{natbib}

\usepackage{amsmath}
\usepackage{color}
\usepackage{multicol}
\usepackage{multirow}

\usepackage[tableposition=top]{caption}
\usepackage{float}
\usepackage{subfigure}
\usepackage{epsfig}
\usepackage{graphicx}\graphicspath{{./figure/}}

\usepackage[table,xcdraw]{xcolor}
\usepackage{manyfoot}
\usepackage{hyperref}
\usepackage{url}
\hypersetup{urlcolor=blue}

\newfootnote{A}
\newcounter{footnoteA}

\newcommand{\myUrl}[1]{%
    \footnote{\url{#1}}}
               
\begin{document}

\title{Convolutional Neural Network-Based Image 
Representation for Visual Loop Closure Detection}




%
\author{\authorblockN{Yi Hou\authorrefmark{1}\authorrefmark{2}, Hong Zhang\authorrefmark{2} and Shilin Zhou\authorrefmark{1}}
\authorblockA{\authorrefmark{1}
College of Electronic Science and Engineering, National University of Defense Technology, Changsha, Hunan, P. R. China\\}
\authorblockA{\authorrefmark{2}
Department of Comput­ing Science, University of Alberta, Edmonton, Alberta Canada\\
Email: {\{yihouhowie, slzhoumailbox\}}@gmail.com, hzhang@ualberta.ca}}

\maketitle

\begin{abstract} 
%
%
Deep convolutional neural networks (CNN) have recently been shown in many 
computer vision and pattern recognition 
applications to outperform by a significant margin state-of-the-art 
solutions that use traditional hand-crafted features. 
However, this impressive performance is yet to be fully exploited in robotics. 
In this paper, we focus one specific problem that can benefit from the recent development of the 
CNN technology, i.e., we focus on using a pre-trained CNN model as a method
of generating an image representation appropriate for {\it visual} loop closure 
detection in SLAM (simultaneous localization and mapping).   
We perform a comprehensive evaluation of the outputs at the intermediate layers of a CNN as image descriptors, in comparison with state-of-the-art image descriptors,
in terms of their ability to match images for detecting loop closures.
The main conclusions of our study include: (a) CNN-based image representations
perform comparably to state-of-the-art hand-crafted competitors in environments
without significant lighting change, (b) they outperform state-of-the-art 
competitors when lighting changes significantly, and (c) they are  
also significantly faster to extract than the state-of-the-art hand-crafted features even on a conventional CPU and are two orders of magnitude faster on an entry-level GPU.
\end{abstract}

\IEEEpeerreviewmaketitle

\section{Introduction}
%
%
Loop closure detection is considered one of the most important problems in 
SLAM (simultaneous localization and mapping). A SLAM algorithm 
aims to map an unknown environment while simultaneously localizing the robot~\cite{Durrant-Whyte2006}.  
Loop closure detection is the problem of determining whether a mobile robot has returned to a previously visited location, and it 
is critical for building a consistent map of the environment by correcting errors that accumulate over time.  
In this paper we are interested in visual loop closure detection, which formulates a solution to the problem by using visual data, i.e., using images captured by the robot.  Despite significant progress in visual loop closure detection, challenges remain especially in dynamic environments that experience, for example, changing illumination conditions.
%
%
\\
\indent
To develop a loop closure detection algorithm, one class of popular and 
successful techniques is based on matching the current view of the robot 
with those in the robot map that correspond to previously visited locations.  In this case, the problem of loop closure detection is essentially one of image matching.  Image matching typically proceeds in two steps: image description and similarity measurement.  An image descriptor compresses an image into a one-dimensional vector that is more compact and discriminating than the original image, and it is the most critical step in visual loop closure detection, as well as the focus of this paper. 
%
%
\\
\indent
Many image description techniques exist for visual loop closure detection, and they have enjoyed tremendous success.  However, all these techniques exclusively use \textit{hand-crafted features}, i.e., they are designed through the process of feature engineering where human expertise and insights dominate the development process to achieve the desired characteristics.  For instance, Bag-of-Visual-Words (BoVW) \cite{Sivic2003BOVW} is introduced in FAB-MAP \cite{Cummins2008aFABMAP, CumminRSS09FABMAP}. Because of the invariance properties of local image features such as SIFT and SURF in building the BoVW descriptor, FAB-MAP achieved an excellent performance, becoming one of the standard baseline algorithms in loop closure detection research. Rather than using local features, GIST is a low dimensional global image descriptor that uses a Gabor filter bank to produce~\cite{Oliva2001GIST}, and it has been popular in recent visual SLAM research~\cite{singh2010visualWhole, sunderhauf2011briefWhole, ZhangYang2012GIST}.
%
%
Image descriptors based on hand-crafted features often share common weaknesses including their lack of robustness with respect to illumination changes and high computational cost. 
\\
\indent
Recent advances in deep learning and convolutional neural networks~\cite{NIPS2012CNN} motivate us to investigate CNN as a potential solution to the weaknesses in existing image descriptors.  In numerous studies, the ability of CNN to learn from visual data features of increasing levels of abstraction has led to its dominating performance over solutions that use hand-crafted features on various standard computer vision benchmarks~\cite{NIPS2012CNN, ChatfieldSVZ14, SimonyanZ14aVeryDeepCNN, Wan2014CBIR, Babenko2014}. 
In particular, the remarkable achievements of CNN on image classification~\cite{NIPS2012CNN, ChatfieldSVZ14} and image retrieval tasks~\cite{Wan2014CBIR, Babenko2014} are extremely encouraging. Considering that visual loop closure detection is similar in spirit to image classification and image retrieval, it is reasonable to expect that the power of CNN-based features can be leveraged in devising a solution to the problem of visual loop closure detection.
%
%
At the same time, it is important to understand the differences between image classification, image retrieval, and loop close detection.  Image classification deals with the problem of categorizing a query image into one of a finite number of known classes; image retrieval is similar to image classification and attempts to find the most similar images of the same class in a database.  In contrast, visual loop closure detection looks for identical images to the current view, which in general includes objects of many classes and may experience dynamic objects and lighting variation.   As a result, an image representation that works well for image classification may not work well for visual loop closure detection and vice versa.  CNN models trained for image classification and image retrieval typically include several fully-connected final layers after the convolutional layers and before the classification step. Interestingly, as we will show later in the paper, features from these fully connected layers do not work well for visual loop closure detection because of the loss of spatial information.  Instead, we need to examine intermediate layers of a deep convolutional neural network for best representations for our application.  This is an open issue to be resolved, one that we attempt to answer in this paper.  
%
%
\\
\indent
Our study uses a publically available, pre-trained CNN model, trained on the a scene-centric dataset called Places \cite{NIPS2014Places} with over 2.5 million images of 205 scene categories.  The pre-trained model serves as an efficient whole-image descriptor generator.  With the trained model, we are able to extract CNN whole-image descriptors easily, one from each layer.  The further and deeper into the CNN pipeline, the more abstract the representation of the input image.  At the same time, the length of the descriptors also varies with the layer depth, representing different computational costs when used for matching images. 
The fact that CNN is able to learn high-level abstractions seems to suggest 
that it is able to exract semantic information that is difficult to obtain
with hand-crafted features, and this leads one to believe 
that a CNN-based image descriptor can exhibit a higher degree of 
invariance properties. 

To evaluate which layer of a CNN is the most appropriate for 
visual loop closure detection, we compare the performance of 
CNN-generated descriptors from all layers of the pre-trained CNN model. 
Based on the result of this evaluation, we select the most promising 
ones and compare them with state-of-the-art hand-crafted image
descriptors in the application of visual loop closure detection, using 
datasets that involve either constant or varying lighting conditions. 
As we will see that the best performing CNN-based image descriptors achieve
a comparable performance to best-performing hand-crafted 
descriptors in environments that 
experience little illumination change, but significantly 
outperform hand-crafted descriptors in environments that undergo lighting 
changes.  Furthermore, comparison in terms of computational cost
shows that CNN-based image descriptors are superior to 
hand-crafted descriptors by one order of magnitude on a CPU and by two
orders of magnitude on an entry-level GPU.
To the best of our knowledge, this is the first time that the performance 
of a CNN-based solution has been comprehensively compared with state-of-the-art
image descriptors in the application of visual loop closure detection.  \\
\indent
%
%
The rest of this paper is organized as follows. Section II gives a brief 
introduction to the related work on visual features for loop closure 
detection, from shallow hand-crafted ones to the ones learned with deep neural networks.  In Section III we present the details of the pre-trained CNN
model and how it is used to generate image descriptors. 
Section IV shows experimental results on three datasets to 
compare the performance of various competing image descriptors. Finally, 
we conclude the paper in Section V with a short discussion and future work.

\section{Related Work}

As mentioned above, our focus in this paper is to  investigate one aspect of visual loop closure detection, namely, image description. In this section, we first give a brief review of four shallow hand-crafted descriptors: BoVW and GIST -- which are popular in visual loop closure detection -- and VLAD and Fisher vector, which are global image descriptors that enjoy the best performance in image classification and retrieval applictions.  We then outline several representative 
computer vision applications -- such as image 
classification and image retrieval -- where CNN has been highly successful.

\subsection{Shallow Hand-Crafted Features}
In the literature, hand-crafted visual image descriptors fall into two 
categories: those that are based local keypoint descriptors and those
that are computed for the entire image.
The BoVW descriptor, arguably the most successful image descriptor in 
visual loop closure, is based on local keypoint descriptors.  
Originally proposed for image retrieval \cite{Sivic2003BOVW}, 
BoVW characterizes an image as a histogram of visual words where visual words are simply vector-quantized versions of the local keypoint descriptors such as SIFT \cite{Lowe2004SIFT} and SURF \cite{Bay2006SURF}. To build a BoVW descriptor, a visual vocabulary is first created offline by clustering a large number of keypoint descriptors whose cluster centres form the visual words of the vocabulary.  In online operation, the local keypoints of a given image are first detected and described.  Each descriptor is then vector-quantized and the histogram of the vector-quantized keypoint descriptors is used as the image descriptor.  BoVW is adopted successfully in the FAB-MAP~\cite{Cummins2008aFABMAP} for computing image similarity and has become one of the most popular techniques in visual loop 
closure detection algorithms.
\\
\indent
Extending the basic idea of BoVW, two other state-of-the-art hand-crafted image 
descriptors based on local visual features of an image are widely used.
Fisher vector (FV)~\cite{Perronnin2007FV, Perronnin2010IFV}, reported as the best hand-crafted image descriptor according to some~\cite{Chatfield11, ChatfieldSVZ14}, uses a Gaussian mixture model (GMM) to construct a visual word dictionary where the means of the Gaussian components are cluster centres as in BoVW and the covariance captures the distribution of the keypoint descriptors within the cluster.  
For each local keypoint descriptor of an image to be described, FV performs
a soft assignment to {\it all} the clusters or Gaussian components and uses the sum of the weights from this assignment to create the final representation.  FV therefore uses second-order statistics, in constrast to the hard assignment of each keypoint descriptor to a single visual word in BoVW. 
Because FV encodes richer information than BoVW, it has been shown to outperform BoVW with in image classification or other visual tasks~\cite{Sanchez2013}. 

Compared to the FV, vector of locally aggregated descriptors (VLAD) descsriptor~\cite{Jegou2010VLAD, Arandjelovic2013IVLAD}, is preferable when the trade-off between the performance and memory footprint of an image descriptor is important. The VLAD is regarded as a simplification of the FV \cite{Jegou2012}. Different from the FV, the VLAD only considers the means so that it uses the first order statistical information of the train keypoint descriptors.  In spite of this, the VLAD is almost comparable in performance to FV in some 
cases~\cite{Arandjelovic2013IVLAD}.
\\
%

Obviously, the performance of the above hand-crafted image descriptors depends on that of local keypoint descriptors, which are developed to provide limited
invariance to affine transformation and illumination. In addition, the hand-crafted image descriptors can be time-consuming to extract because of the keypont
detection process and their vector quantization.
To overcome these weaknesses, GIST, a whole-image descriptor~\cite{Oliva2001GIST}, has been studied in recent visual loop closure detection research~\cite{singh2010visualWhole, sunderhauf2011briefWhole, ZhangYang2012GIST}. The GIST descriptor is generated by measuring the responses of an image to a Gabor filter bank and it is compact, with less than 1000 dimensions in its standard implementation. 
However, since the GIST descriptor is computed for an entire image, its limited robustness with respect to image transformations such as camera motion and illumination variation may hamper its effectiveness in image matching applications. 

\subsection{Deep CNN-Based Features}

One prominent area of CNN research is centered around the problem of 
image classification \cite{NIPS2012CNN}. CNN's breakthrough performance on the ImageNet LSVRC-2010 demonstrated the power of deep learning. Recently, a comprehensive evaluation further demonstrated the advantages of deep CNN features with respect to shallow hand-crafted features for image classification~\cite{ChatfieldSVZ14}. Through empirical experiments, the fully-connected 7 (FC7) features were used and achieved better results than all hand-crafted features. 
Besides image classification, CNN features have also become the winners in image retrieval benchmarks. In \cite{Wan2014CBIR}, a framework with several CNN feature generalization schemes for content-based image retrieval was proposed. A similar study was presented in \cite{Babenko2014}, although it placed the emphasis on investigating the selection of the best layer of CNN features and the effect of PCA compression.  The experimental results in~\cite{Wan2014CBIR, Babenko2014} show that the CNN-based deep features generated at the full-connected layers achieve much better performance than hand-crafted features in the image retrieval application.  As mentioned in the introduction, this conclusion may be not directly applicable to the problem of visual loop closure detection. In spite of this, these exciting achievements still have convinced us that a new door for visual loop closure detection studies has been opened and that we can develop new effective solutions by taking advantage of the power of CNN in abstracting an image with semantic information.

\section{CNN-based Image Descriptors}
  
\subsection{CNN Architecture}

In our study, we use Caffe \cite{jia2014caffe}, which is a open-source deep learning framework, to extract CNN-based features.
The architecture of a standard CNN \cite{NIPS2012CNN}, which is reconstructed in Caffe, is briefly summarized in Table \ref{tab:ArchitectureCNN}. 
This CNN model is a multi-layer neural network that mainly consists of three types of layers: five convolutional layers, three max-pooling layers and three fully-connected layers. Note that a max-pooling layer follows the first, second and fifth convolutional layer but not the third and fourth convolutional layers. A max pooling layer provides translation invariance to the correspoinding features and reduce their dimensions at the same time. In fact, it is also a process of building an abstract representation, by merging lower-level local information. This abstraction occurs locally within a neighborhood window.  In contrast, for a fully-connected layer, all neurons in the previous layer are fully connected to every single neuron of the current layer. This connection can be an important step in high-level reasoning that is beneficial to image classification and image retrieval applications. However, spatial information of an image is lost through a full-connected layer, and this may not be desirable in applications such as visual loop closure detection.
\\
\indent
With a deep architecture, CNN is able to learn high-level semantic features at various levels of abstraction. Further we can expect that a deeper pooling layer such as Pool 5, to be particularly promising for visual loop closure detection since it still retains much of the spatial information of the input image and derives richer semantic representations of an input image than shalower convolutional and pooling layers. 

\begin{table*}[tp]\renewcommand{\arraystretch}{1.2} \footnotesize
	\setlength{\abovecaptionskip}{3pt}
	\caption{Architecture of the standard CNN model in Caffe and the dimension of the CNN-based feature at each layer.}
    \centering
	\begin{tabular}{c||c|c|c|c|c|c|c|c||c|c|c}
	\hline \hline
	\multirow{3}{*}{\textbf{Layer}} & \multicolumn{8}{c||}{\textbf{Convolutional}} & \multicolumn{3}{c}{\textbf{Fully-Connected}} \\
	\cline{2-12}
	 & CONV1 &  & CONV2 &  & CONV3 & CONV4 & CONV5 &  & FC6 & FC7 & FC8 \\ \cline{2-12} 
 &  & POOL1 &  & POOL2 &  &  &  & \textbf{POOL5} &  &  &  \\
 	\hline
 	\multicolumn{1}{c||}{\textbf{Dimension}} & \multicolumn{1}{c|}{290400} & \multicolumn{1}{c|}{69984} & \multicolumn{1}{c|}{186624} & \multicolumn{1}{c|}{43264} & \multicolumn{1}{c|}{64896} & 64896 & 43264 & \textbf{9216} & 4096 & 4096 & 1000 \\ 
 	\hline \hline
 	\end{tabular}
 	\label{tab:ArchitectureCNN}
    \setlength{\belowcaptionskip}{-3pt}
\end{table*}

\subsection{CNN Whole-Image Descriptors}

Using a pre-trained standard CNN model, we can create whole-image descriptors, one from each layer, by traveling along the depth of the network.
That is, as an input image is passed through a CNN, the output of each layer is considered as a feature vector, $X$.  We then perform the important step of normalization on $X$~\cite{NIPS2012CNN, ChatfieldSVZ14}, using for example ${\ell}_2$-norm as follows:
\begin{equation}\label{equ:L2Normalization}
  (x_1, ..., x_d) \leftarrow (\dfrac{x_1}{\sqrt{\sum_{i=1}^{d}{x_i}^{2}}}, ..., \dfrac{x_d}{\sqrt{\sum_{i=1}^{d}{x_i}^{2}}})
\end{equation}
where $(x_1, ..., x_d)$ is a feature vector $X$ with $d$ dimensions.
\\
\indent
In this way we can create multiple CNN whole-image descriptors, which are layer-by-layer abstract representations of the input image.  These high-level CNN-based descriptors are tested for their performance in visual loop closure detection. Note that in Table \ref{tab:ArchitectureCNN}, we also list the dimensions of the CNN-based image descriptors according to the parameter settings of the standard CNN model in Caffe.


\section{Experimental Results}

\subsection{Datasets and Ground Truth Generation}

Experiments are conducted on three datasets, two of which publicly available and one created by ourselves for the purpose of examining the illumination invariance property.
{\it City Centre} and {\it New College} are two datasets widely used in visual SLAM research and in evaluating loop closure detection in particular.  
They are published in \cite{Cummins2008aFABMAP}, and are used in our 
experiments.  The two 
datasets contain 1237 and 1073 image pairs respectively, collected by a mobile robot with two cameras on the left and right side when it is driven through an outdoor urban environment with stable lighting conditions.  Ground truths in terms of true loop closures are also available.  Details of these two datasets are available online \myUrl{http://www.robots.ox.ac.uk/~mobile/IJRR_2008_Dataset/} .

{\it UA Campus} is a dataset that we created on the campus of University of America\footnote{real name changed in order to be anonymous} in order to
evaluate the performance of image descriptors in the case changing lighting conditions.  The robot in this dataset has a forward facing camera and covers a route of about 650 meters.  In order to create illumination variation, the robot was driven through the same route at five different times of the day (specifically starting at 06:20, 10:05, 14:10, 16:40 and 22:15, respectively).  Thus, five subsets have been acquired.  The images in the five sequences are then manually matched to generate loop closure ground truth. 
Details of five datasets are described in Table \ref{tab:DetailsDatasetCS} and their example images acquired at the same location are shown in Fig. \ref{fig:ExampleImgCS}. For convenience, five subsets are named as 0620, 1005, 1410, 1640 and 2215.
%
%
\begin{table}[tp]\renewcommand{\arraystretch}{1.2}\addtolength{\tabcolsep}{-3pt} \footnotesize
	\setlength{\abovecaptionskip}{3pt} 
	\caption{Details of UA Campus dataset which consists of five subsets acquired at five different times.}
    \centering
    \begin{tabular}{c|c|c|c|c|c}
    \hline\hline
    Subset & 0620 & 1005 & 1410 & 1640 & 2215    \\ \hline
    Feature & dawn & overcast AM & \begin{tabular}[c]{@{}c@{}}rainy PM\end{tabular} & \begin{tabular}[c]{@{}c@{}}sunny PM\end{tabular}  & dusk\\
    \hline\hline
    \end{tabular}
    \label{tab:DetailsDatasetCS}
    \setlength{\belowcaptionskip}{-3pt} 
\end{table}
%
%
\begin{figure*}[tp] \footnotesize
  \centering
  \setlength{\abovecaptionskip}{0pt}
  \subfigure[0620]{
  \includegraphics[width=0.15\textwidth]{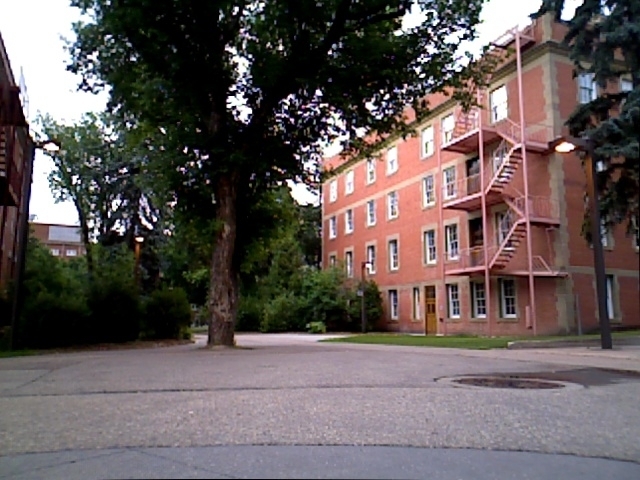}
  \label{ExampleImgCS-0620}
  }
  \hspace{.03in}
  \subfigure[1005]{
  \includegraphics[width=0.15\textwidth]{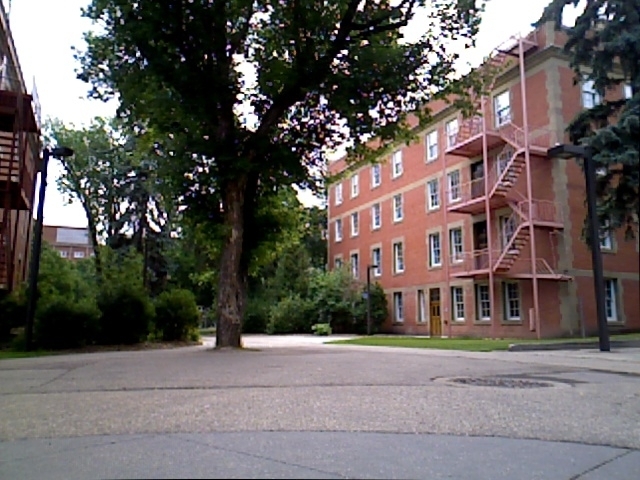}
  \label{ExampleImgCS-1005}
  }
  \hspace{.03in}
  \subfigure[1410]{
  \includegraphics[width=0.15\textwidth]{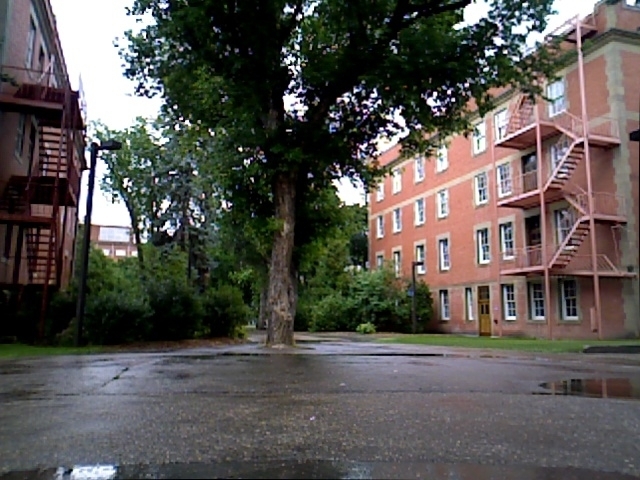}
  \label{ExampleImgCS-1410}
  }
  \hspace{.03in}
  \subfigure[1640]{
  \includegraphics[width=0.15\textwidth]{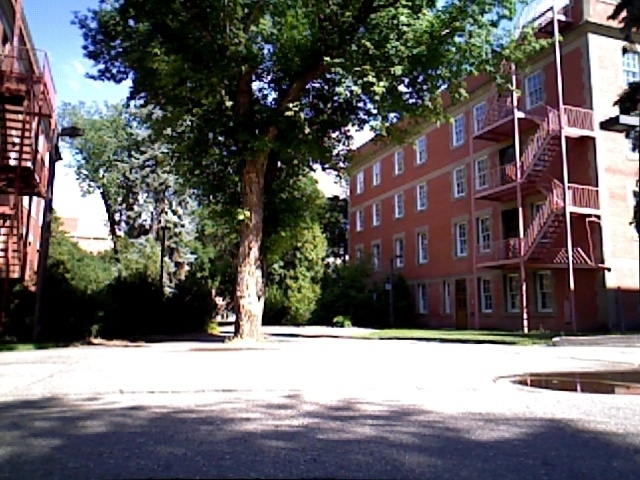}
  \label{ExampleImgCS-1640}
  }
  \hspace{.03in}
  \subfigure[2215]{
  \includegraphics[width=0.15\textwidth]{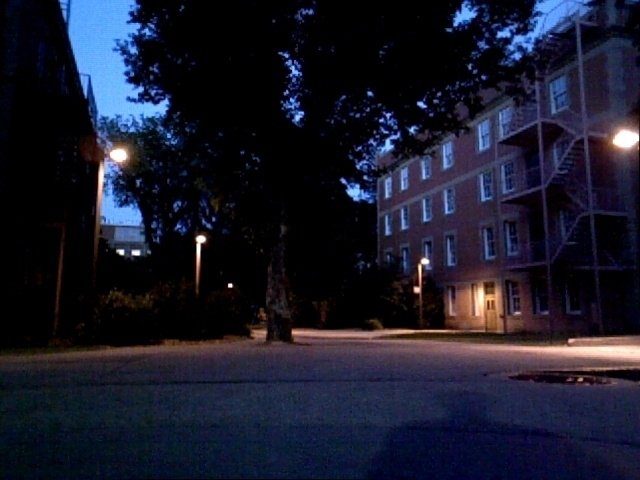}
  \label{ExampleImgCS-2215}
  }
  \caption{Five example images of a same location corresponding to five subsets in UA Campus dataset.}
  \setlength{\belowcaptionskip}{-3pt}
  \label{fig:ExampleImgCS}
\end{figure*}

\subsection{Algorithm Implementation Details}
%
%
For hand-crafted features, we include BoVW, FV and VLAD under the implementations provided by the VLFeat toolbox~\cite{Vedaldi2010VLFeat}, which is an popular open source library of computer vision algorithms. The GIST descriptor is computed by using its open code \myUrl{http://people.csail.mit.edu/torralba/code/spatialenvelope/}. 
%
%
For learned deep CNN features, we use Caffe~\cite{jia2014caffe} with a reconstruction of the standard CNN architecture~\cite{NIPS2012CNN} to extract CNN-based image descriptors.
%
%

In hand-crafted features, we kept the default parameter settings for GIST~\cite{Oliva2001GIST}.  For BoVW, a single dictionary with 1024 visual words is 
built from the 128 dimensional SIFT descriptors in training images. For FV and VLAD, we use the suggested settings~\cite{Chatfield11}: the dimension of SIFT descriptors are reduced from 128 dimensions to 80 dimensions by PCA, and then GMMs with K = 256 components are learned from these SIFT descriptors. No spatial scheme is used because we do not use augmentation for computing CNN-based features. Thus the feature dimensionalities of GIST, BoVW, FV and VLAD are 512, 1024, 40960 and 20480, respectively. 
\\
\indent
For CNN-based features, the default settings are used except for previous mentioned augmentation strategy. Note that we use the Places-CNN model \cite{NIPS2014Places} in our experiments, which shares the same CNN architecture with the standard ImageNet-CNN model \cite{NIPS2012CNN} but is trained on a scene-centric database rather than the ImageNet database. The model is better suited for scene recognition and loop closure detection than the CNN model trained on ImageNet.
%
%
\\
\indent
We use two criteria for performance evaluation: precision-recall curve and average precision. Precision-recall curve is a standard method of evaluation in pattern recognition in general and in loop closure detection in particular.  To produce the precision-recall curve of a given image descriptor in our case, we compute for each of the query images (current robot views) its descriptor and then find its nearest neighbor in the robot map according to Euclidean distance.  A threshold on the distance (which measures similarity) is then applied to determine if loop closure has occurred, and a precision and recall pair results after all images in the dataset are considered.  By varying the distance threshold, we can then produce a precision-recall curve.  The second criterion, average precsion, is useful when we want a scalor value to characterize the overall performance of loop closure detection.  In general, a high precision over all recall values is desirable, and average precision captures this property by computing the average of the precisions over all recall values of a precision-recall curve. 

For the City Centre and New College datasets, we evaluate the image descriptors for visual loop closure detection in a straightforward way. For the UA Campus dataset, which has five subsets for five different times of a day, we treat the first sbuset, 0620, as the reference run, and each of the other four subsets as revisits of this same route in 0620. Therefore, we can perform four experiments of visual loop closure detection on the UA Campus, which are labeled as \textit{vs1005}, \textit{vs1410}, \textit{vs1640} and \textit{vs2215} in the experimental results we will see shortly.

%
%
\begin{table*}[tp]\renewcommand{\arraystretch}{1.2}\addtolength{\tabcolsep}{-3pt} \footnotesize
	\setlength{\abovecaptionskip}{3pt} 
	\caption{Average Precisions of CNN-based features and hand-crafted features on the City Centre and New College datasets. (The representatives of CNN layers chose in Section \ref{SectionIVC} are bold.)
	}
    \centering
    \begin{tabular}{c||c|c|c|c||c|c|c|c|c|c|c|c|c|c}
    \hline\hline
    \multirow{2}{*}{Feature} & \multicolumn{4}{c||}{Hand-crafted} & \multicolumn{10}{c}{CNN} \\
    \cline{2-15} 
	  &GIST &BoVW &FV &VLAD &POOL1 &CONV2 &POOL2 &\textbf{CONV3} &CONV4 &CONV5 &\textbf{POOL5} &\textbf{FC6} &FC7 &FC8 \\ \hline\hline
    City Centre &0.81021 &0.81246 &0.85176 &0.87275 &0.70866 &0.74749 &0.8027 &\textbf{0.80365} &0.81753 &0.82593 &\textbf{0.84797} &\textbf{0.79774} &0.7447 &0.57118 \\ \hline
	New College &0.74813 &0.79761 &0.93055 &0.93903 &0.58493 &0.67063 &0.72377 &\textbf{0.73614} &0.75718 &0.77679 &\textbf{0.82422} &\textbf{0.81671} &0.76652 &0.61788 \\
    \hline\hline
    \end{tabular}
    \label{tab:APCityCentreNewCollege}
    \setlength{\belowcaptionskip}{-3pt} 
\end{table*}

\begin{figure*}[tp] \footnotesize
  \centering
  \setlength{\abovecaptionskip}{0pt}
  \subfigure[Comparison of CNN-based features]{
  \includegraphics[width=0.45\textwidth]{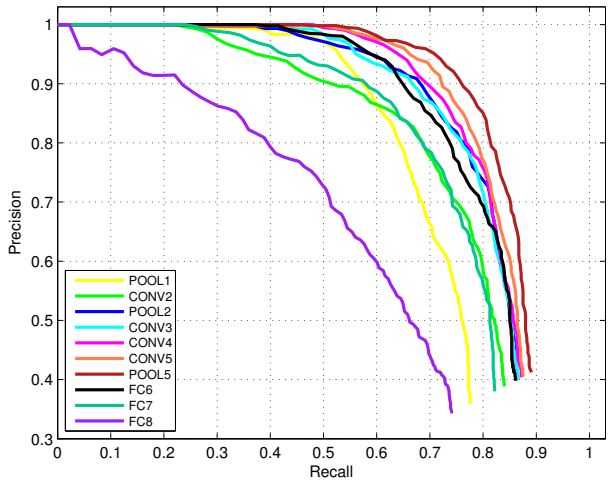}
  \label{PRCityCentreCNNvsCNN}
  }
  \subfigure[Comparison of hand-crafted and CNN-based features]{
  \includegraphics[width=0.45\textwidth]{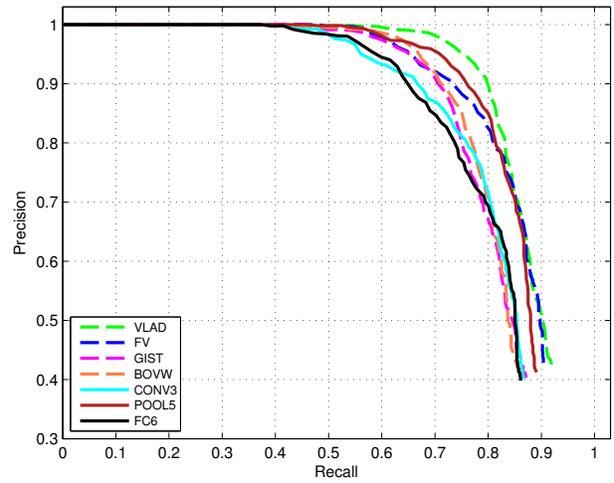}
  \label{PRCityCentreCNNvsHand}
  }
  \caption{Precision-Recalls on the City Centre dataset: (a) comparison of CNN-based features, (b) comparison of hand-crafted and CNN-based features.
  }
  \setlength{\belowcaptionskip}{-3pt}
  \label{fig:PRCityCentreCNN}
\end{figure*}

\begin{figure*}[tp] \footnotesize
  \centering
  \setlength{\abovecaptionskip}{0pt}
  \subfigure[Comparison of CNN-based features]{
  \includegraphics[width=0.9\columnwidth]{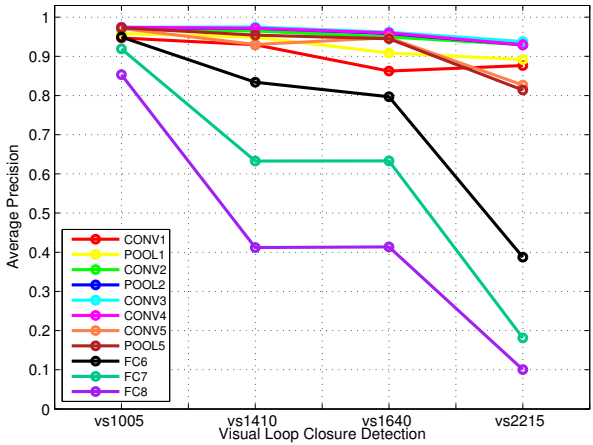}
  \label{APCompSciCNNvsCNN}
  }
  \subfigure[Comparison of hand-crafted and CNN-based features]{  
  \includegraphics[width=0.9\columnwidth]{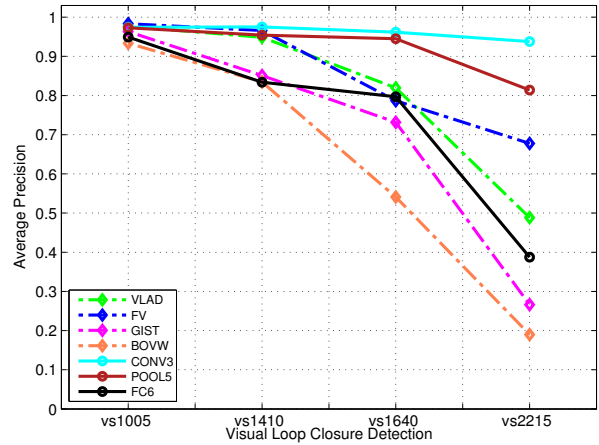}
  \label{APCompSciCNNvsHand}
  }
  \caption{Average Precisions on the UA Campus dataset: (a) comparison of CNN-based features, (b) comparison of hand-crafted and CNN-based features.}
  \setlength{\belowcaptionskip}{-3pt}
  \label{fig:APCompSci}
\end{figure*}

\begin{figure*}[tp] \footnotesize
  \centering
  \setlength{\abovecaptionskip}{0pt}
  %
  %
  \subfigure[ Comparison of CNN-based features: vs1005]{
  \includegraphics[width=0.45\textwidth]{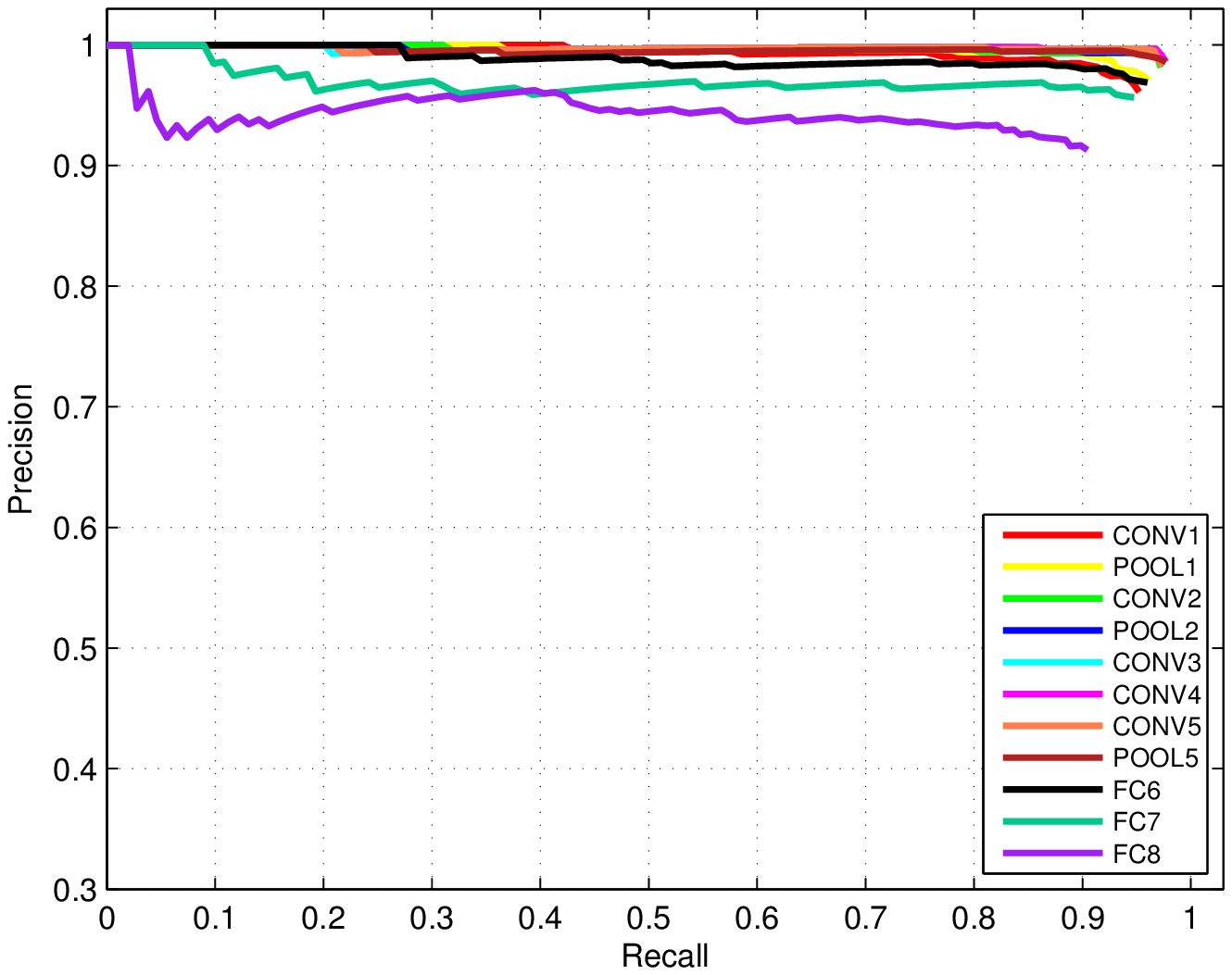}
  \label{PRCompSciCNNvsCNN-0620vs1005}
  }
  \subfigure[ Comparison of CNN-based features: vs2215]{
  \includegraphics[width=0.45\textwidth]{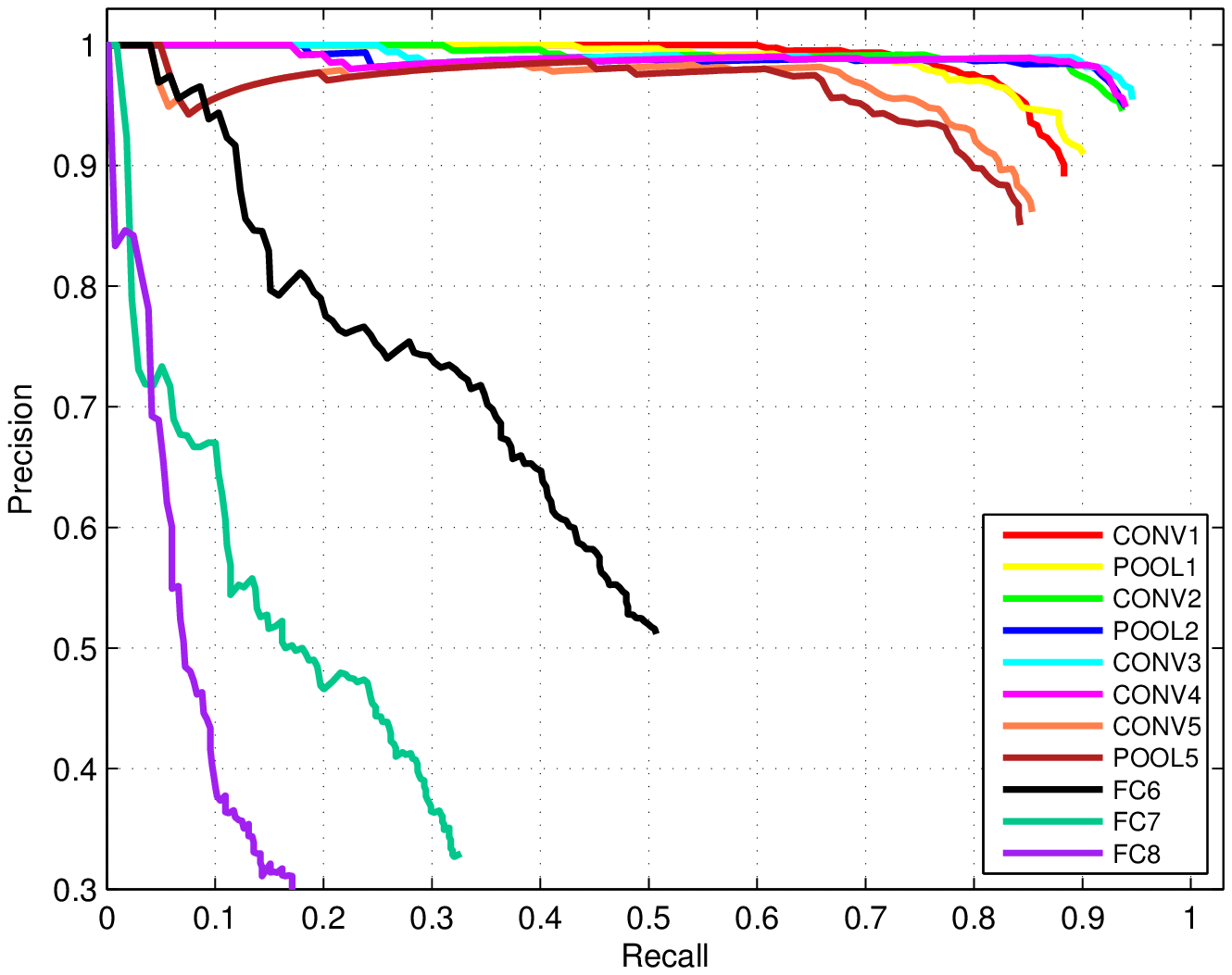}
  \label{PRCompSciCNNvsCNN-0620vs2215}
  }
  %
  %
  \subfigure[Comparison of hand-crafted and CNN-based features: vs1005]{
  \includegraphics[width=0.45\textwidth]{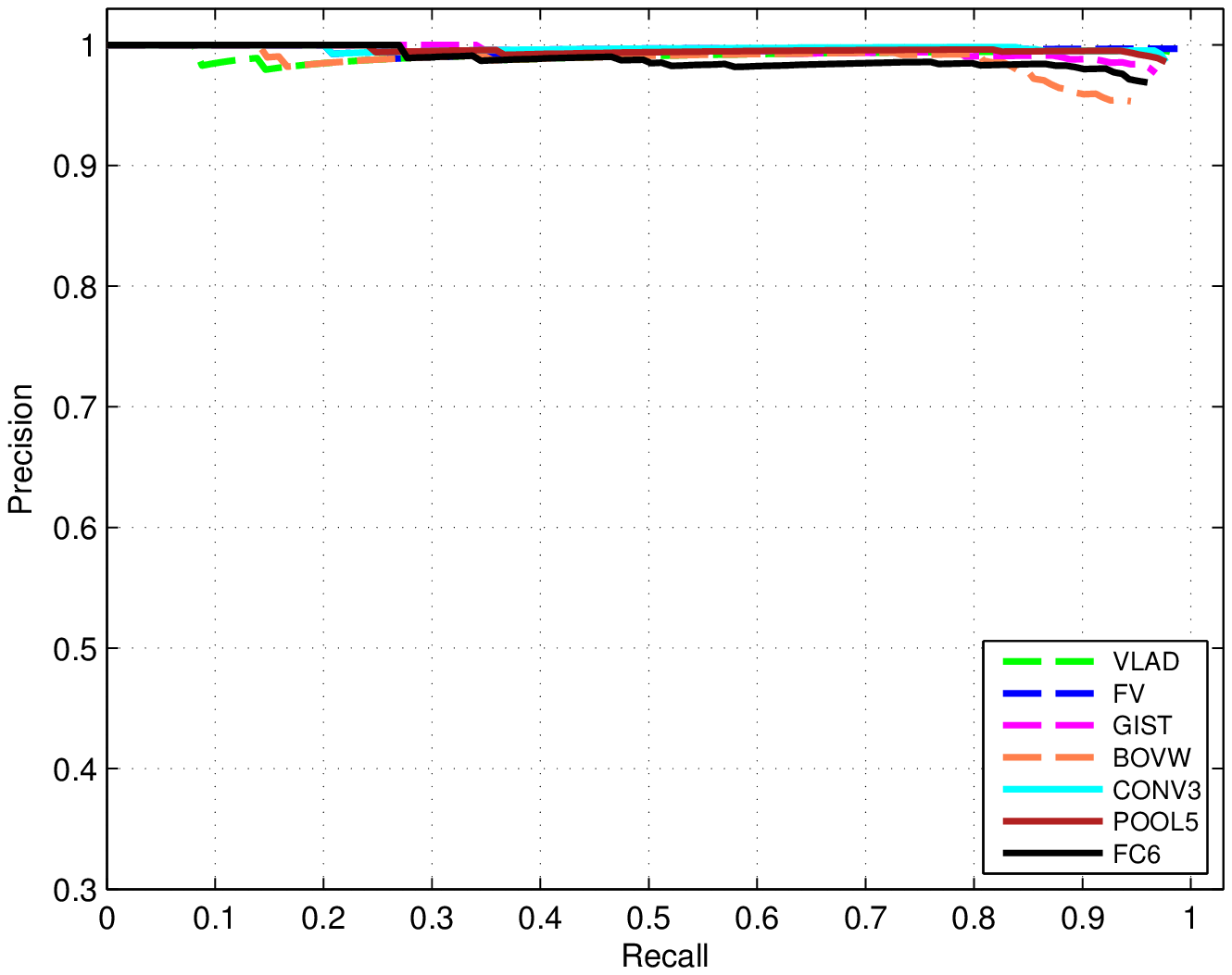}
  \label{PRCompSciCNNvsHand-0620vs1005}
  }
  \subfigure[Comparison of hand-crafted and CNN-based features: vs2215]{
  \includegraphics[width=0.45\textwidth]{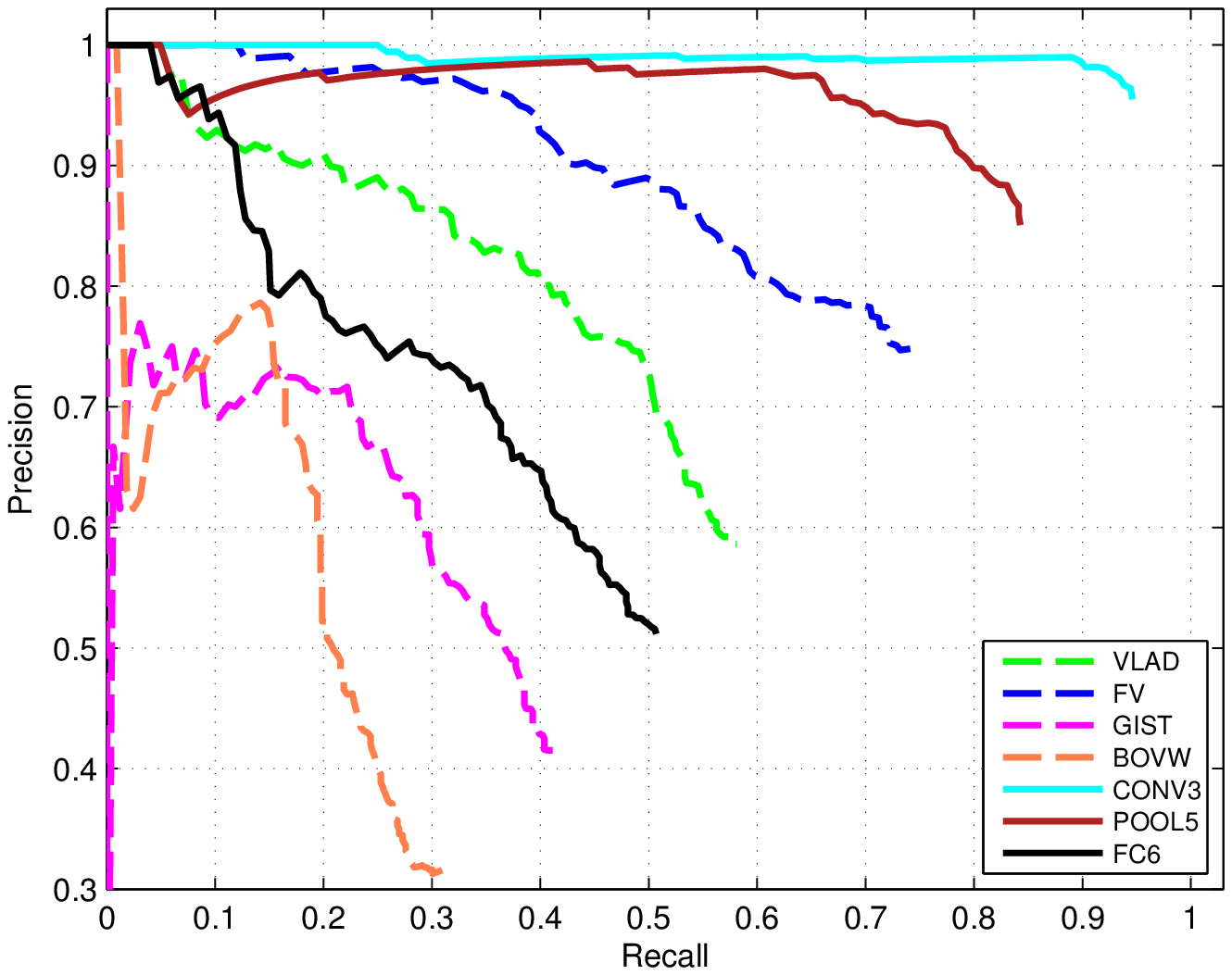}
  \label{PRCompSciCNNvsHand-0620vs2215}
  }
  \caption{Precision-Recalls on two visual loop closure detection (vs1005 and vs2215) from the UA Campus dataset: (a) and (b) comparison of CNN-based features, (c) and (d) comparison of hand-crafted and CNN-based features.}
  \setlength{\belowcaptionskip}{-3pt}
  \label{fig:PRCompSciCNNvsCNN}
\end{figure*}

\subsection{Comparison of CNN-Based Descriptors}
\label{SectionIVC}

To find out the most applicable CNN-based features for visual loop closure detection, we evaluate the performance of CNN-based features from all layers (except for the CONV1 layer) under similar and changing lighting conditions. 
Experimental results on the City Centre and New College datasets, in terms of precision-recall curves and average precision are shown in Fig. \ref{PRCityCentreCNNvsCNN} and Table \ref{tab:APCityCentreNewCollege}, respectively. 
For the convolutional and max-pooling layers, the performances are increasing layer by layer until the final fully-connected layers, indicating the importance of depth in deep learning and that of spatial information when encoding an image. Based on these results and conditional on our experimental settings, we can draw the following conclusions:
\begin{itemize}
\item 
POOL5 of the CNN achieves the best result over all layers, 
\item
FC6 is the best among three fully-connected layers, 
and 
\item
the performances of the last three convolutional layers are comparable.
\end{itemize}

When lighting change is introduced, as is present in
the UA Campus dataset, performances of the competing CNN descriptors in terms of the average precision and precision-recall curve are shown in Figs. \ref{APCompSciCNNvsCNN}, \ref{PRCompSciCNNvsCNN-0620vs1005} and \ref{PRCompSciCNNvsCNN-0620vs2215}. 
Among the convolutional layers, all their performances are consistently excellent with or without lighting change.   
So we also select the CONV3 as the representative of the convolutional layers. For the max-pooling layers, although there is a slight decline with the severe lighting changing for the POOL5 layer (see Fig. \ref{PRCompSciCNNvsCNN-0620vs2215}), we will still adopt it as the winning representative of the pooling layers for its minimum dimension and excellent performance in the previous two datasets. FC6 also achieves the best performance among the three fully-connected layers although all fully-connected layers perform considerably worse than both convolutional and pooling layers.  In summary, from the experiments on UA Campus dataset, we can conclude that
\begin{itemize}  
\item although the performances of the convolutional and max-pooling layers are a little better than the fully-connected layers when lighting is similar,
\item they significantly outperform the fully-connected layers when lighting changes dramatically.
\end{itemize}

In view of the all of the results above, we choose the \textbf{CONV3}, \textbf{POOL5} and \textbf{FC6} as the representatives of 11 CNN layers, in our subsequent comparative study with hand-crafted image descriptors. As speculated earlier, CNN features from the fully-connected layers, which typically work well on image classification and image retrieval benchmarks, are not the best performing for visual loop closure detection, especially in case of illumination variation.  If we consider both performance and compactness, the CNN descriptor from POOL5 is arguably the most appropriate for visual loop closure detection especially when large environmental maps with numerous locations are involved.

\subsection{Comparison of Hand-Crafted and CNN Image Descriptors}

To compare the effectiveness of competing image descriptors for visual loop closure detection, we conduct similar experiments as above.  Comparison in this experiment includes three CNN-based descriptors, as justified above, and  four hand-crafted features: BoVW, GIST, FV and VLAD. The experiment is also conducted on the same three datasets, two without illumination change and one with the change. 

For the City Centre and New College datasets, the performance of both CNN-based descriptors from all layers including CONV3, POOL5 and FC6 and of the hand-crafted image descriptors is already presented in Table \ref{tab:APCityCentreNewCollege}.  The precision-recall curves of all seven competing descriptors on the City Centre dataset are shown in Fig.~\ref{PRCityCentreCNNvsHand}. We can see that the performance of the representative CNN descriptors is similar to that of the hand-crafted descriptors according to the two evaluation criteria.

When illumination change happens as in
the UA Campus dataset, the relative performance of the competing descriptors
becomes interesting.  The comparison in terms of precision-recall curve is shown in Fig. \ref{PRCompSciCNNvsHand-0620vs2215}, and the comparison in terms of average precision is shown in Fig. \ref{APCompSciCNNvsHand} (right-most column of AP values).
We can see that the performances of hand-crafted features are very sensitive to lighting changes. However, the performances of CNN-based descriptors are relatively insensitive to light change, except for FC6 although it still outperforms BoVW and GIST descriptors.

\subsection{Computational Time}

Another important consideration in evaluating an image descriptor is its computational efficiency.  This efficiency is measured by both the time it takes to extract a descriptor and the length of the descriptor.  We have conducted a
comparion in the extraction time of the competing descriptors, using
a laptop with a 2.40GHz CPU and 8GB memory. In this case, we use the Matlab wrapper for extracting all hand-crafted features and the Python wrapper for extracting the CNN-based features. We also further test the CNN-based descriptor extraction time with the Caffe C++ code on a desktop PC with a NVIDIA Quadro K4200 GPU with 1344 cores.  This entry-level GPU requires a PCI bus but is low-cost and compact enough to be housed in a mobile robot.
\\
\indent
Average extraction times for all descriptors are listed in Table~\ref{tab:TimePerImg}. The reported time here is the average over 649 images, which only includes the feature extraction time but excludes the time to load the input image or the CNN model. For the CPU-based extraction, CNN descriptors are more efficient than all hand-crafted descriptors, taking 0.155 seconds per image on average, about 3 times faster than the efficient GIST descriptor and 10 times faster than other three state-of-art hand-crafted features.  If CNN descriptors are extracted on our GPU, the average extraction time is reduced to 0.019 seconds per image, approximately two orders of magnitude faster than the state-of-the-art hand-crafted image descriptors.
\begin{table}[tp]\renewcommand{\arraystretch}{1.2} \footnotesize
	\setlength{\abovecaptionskip}{3pt} 
	\caption{Average computational time per image for different feature descriptors.}
    \centering
    \begin{tabular}{c||c|c|c|c||c|c}
    \hline\hline
    \multirow{2}{*}{Feature} & \multicolumn{4}{c||}{Hand-crafted} & \multicolumn{2}{c}{CNN} \\
    \cline{2-7} 
	   & GIST & BoVW & FV & VLAD & CPU & GPU \\ \hline
    Time(s) & 0.589 &  1.815 & 1.829 & 1.148 & 0.155 & 0.019 \\
    \hline\hline
    \end{tabular}
    \label{tab:TimePerImg}
    \setlength{\belowcaptionskip}{-3pt}
\end{table}

\section{Conclusion and Discussion}
We have presented in this paper a comparative study between state-of-the-art hand-crafted image descriptors and CNN-based image descriptors in the visual loop closure detection application.  The CNN-based descriptors are extracted from a pre-trained CNN model that is publicably available and developed for scene classification.
The experimental results on three different datasets allow us to conclude that CNN-based image descriptors perform similarly to hand-crafted descriptors in environments without illumination change, but outperform hand-crafted descriptors by a significant margin when the robot navigation environment experiences inevitable illumination change in long term operations.  In addition, CNN-based image descriptors are faster to extract by an order of magnitude on a CPU or two orders of magnitude by a low-cost, entry-level GPU.
Among the CNN-based image descriptors we observe that POOL5 layer provides the best choice in terms of both detection accuracy and compactness of the representation. 
\indent
Note that our study is still preliminary at this point since there are a number of existing online CNN models that can be explored.   In addition, one can fine-tune an existing CNN -- or train a new CNN model -- for the specific task of loop closure detection.  In our future, we will employ dimensionality reduction techniques to minimize the image descriptor size and increasing its discriminating power.  As well, advanced deep-learning techniques such as auto-encoding will be investigated for their application to visual loop closure detection.

\section*{Acknowledgments}

This work was supported partially by NSERC, the Hunan Provincial Innovation Foundation for Postgraduate (No.CX2014B021), the Fund of Innovation of NUDT Graduate School (No.B140406), the Hunan Provincial Natural Science Foundation of China (No.2015JJ3018) and the China Scholarship Council Scholarship. 
The authors appreciate the helpful comments from reviewers.

\bibliographystyle{plainnat}
\bibliography{ICIA2015howie-references}

\end{document}